# In-depth Assessment of an Interactive Graph-based Approach for the Segmentation for Pancreatic Metastasis in Ultrasound Acquisitions of the Liver with two Specialists in Internal Medicine

Jan Egger, Dieter Schmalstieg
Institute for Computer Graphics and Vision
Graz University of Technology
Graz, Austria

Lucas Bettac, Mark Hänle, Alexander Hann
Dept. of Internal Medicine I
Ulm University
Ulm, Germany

Xiaojun Chen
School of Mechanical Engineering
Shanghai Jiao Tong University
Shanghai, China

Tilmann Gräter
Dept. of Diagnostic and Interventional Radiology
Ulm University
Ulm, Germany

Jan Egger
BioTechMed-Graz
Graz, Austria
egger@tugraz.at

Wolfram G. Zoller, Alexander Hann
Dept. of Internal Medicine and Gastroenterology
Katharinenhospital
Stuttgart, Germany

*Abstract*— The manual outlining of hepatic metastasis in (US) ultrasound acquisitions from patients suffering from pancreatic cancer is common practice. However, such pure manual measurements are often very time consuming, and the results repeatedly differ between the raters. In this contribution, we study the in-depth assessment of an interactive graph-based approach for the segmentation for pancreatic metastasis in US images of the liver with two specialists in Internal Medicine. Thereby, evaluating the approach with over one hundred different acquisitions of metastases. The two physicians or the algorithm had never assessed the acquisitions before the evaluation. In summary, the physicians first performed a pure manual outlining followed by an algorithmic segmentation over one month later. As a result, the experts satisfied in up to ninety percent of algorithmic segmentation results. Furthermore, the algorithmic segmentation was much faster than manual outlining and achieved a median Dice Similarity Coefficient (DSC) of over eighty percent. Ultimately, the algorithm enables a fast and accurate segmentation of liver metastasis in clinical US images, which can support the manual outlining in daily practice.

## I. INTRODUCTION

With a survival rate of less than 7%, five years after diagnosis, pancreatic adenocarcinoma (PDAC) has a poor prognosis [1] and is anticipated to become the second most common cause of cancer death in 2030 [2]. In addition, most of the patients are diagnosed when they have already reached a metastatic stage [3]. However, in recent years, new therapeutic options were introduced prolong survival of patients diagnosed in a metastatic state [4], [5] and second line therapies can also increase the duration of treatment [6]-[9]. Regarding these facts, the neuroendocrine neoplasms (NEN), which is the second most common carcinoma arising from the pancreas, have also a poor prognosis when diagnosed with liver metastasis [10]. The number of staging examinations and chemotherapy cycles of these patients, which are in general performed about every two to three months, is also increasing [11] and in that context ultrasound offers a cheap and fast method to visualize focal liver lesions (FLL). Overall, ultrasound enables a similar sensitivity regarding FLL when compared with Magnetic Resonance Imaging (MRI) and Computed Tomography (CT). Moreover, ultrasound is more widespread than MRI or CT and also included as the imaging modality of choice in the German guideline for follow up of patients after resection of stage I or II colon cancer [12]. Additionally, ultrasound is recommended for the response evaluation of pancreatic cancer patients undergoing palliative treatment by the ESMO-ESDO clinical practice guidelines [13]. However, because of the different appearances of liver metastasis in ultrasound images acquisitions, the interpretation of these have a poor inter-observer agreement [14], [15]. Finally, the manual outlining of liver lesions to assess their size is time- consuming and yields to different results due to the poor inter-observer agreement.

Others working in the segmentation [16]-[21] area of US images, are for example Hao et al. [22], Bahrami et al. [23], Bakas et al. [24], Gatos et al. [25], Jain and Kumar [26], Quan et al. [27] and Ciurte et al. [28]. Hao et al. present a region growing that is performed in a multi-feature vector space. In doing so, they develop three criteria for region growing control: (1.) They use global information instead of local information for the region growing. (2.) They introduce a new idea termed "geographic similarity" to overcome the effects of speckle noise and attenuation artifacts. (3.) They employ an equal opportunity competence criterion to make results independent of processing order.

Bahrami et al. present a boundary delineation for hepatic hemangioma in ultrasound images. Their preprocessing phase includes three main stages: (1.) An image contrast enhancement step, using a so-called Difference of Offset Gaussian (DoOG) approach. (2.) The application of a Canny edge filter. (3.) The application of an adaptive threshold in order to detect the hemangioma ROI.

Finally, a snake algorithm is applied to segment the hemangioma region in the second phase.

Bakas et al. present the semi-automatic segmentation of focal liver lesions in contrast-enhanced ultrasound, which is based on a probabilistic model. Therefore, they propose a two-step method, initialized by a single seed point. In a first step, rectangular force functions are applied to improve the accuracy and computational efficiency of an active ellipse model for approximating the focal liver lesion shape. Afterwards, a probabilistic boundary refinement method is applied to iteratively classify the boundary pixels. Gatos et al. report an automated quantification algorithm for the detection and evaluation of focal liver lesions with contrast-enhanced ultrasound. The lesion detection involves wavelet transform zero crossings utilization as an initialization step to the Markov random field model toward a lesion contour extraction. After the lesion detection across frames, a time intensity curve is computed that provides the contrast agents' behavior at all vascular phases with respect to adjacent parenchyma. From each time intensity curve, eight features were automatically calculated and employed into a support vector machines (SVMs) classification algorithm. Jain and Kumar propose region-difference filters for the segmentation of liver ultrasound images. The region-difference filters evaluate the maximum difference of the average of two regions of the window around the center pixel, which results for a whole image in a region-difference image. Afterwards, the region-difference image was converted into a binary image and morphologically operated for segmenting the desired lesion from the ultrasound image. Quan et al. present the segmentation of tumor ultrasound images via a region-based normalized cut (Ncut) method. In a first step, they use a linear iterative clustering algorithm to divide the image into a number of homogeneous over-segmented regions. Subsequently, these regions are interpreted as nodes and a similarity matrix is constructed by comparing the histograms of each two regions. In a last step, the Ncut method is applied to merge the over-segmented regions. Ciurte et al. propose a semi-supervised segmentation approach of ultrasound images based on patch representation and continuous Min Cut [29], [30]. Summarized, they use a graph of image patches to represent the ultrasound image and a user-assisted initialization with labels that acts as soft priors and formulate it as a continuous minimum cut problem solved with an optimization algorithm.

II. METHODS

*A. Data Acquisition*

All images were acquired using a multifrequency-curved probe, which allows ultrasound acquisitions with a bandwith of 1 to 6 MHz (LOGIQ E9/GE Healthcare, Milwaukee, Il, USA and Toshiba Aplio 80, Otawara, Japan). Images of liver metastasis were selected retrospectively from patients with PDAC or NEN in the digital picture archive of the ultrasound unit of the Katharinenhospital Stuttgart (Germany).

*B. Measurement*

Manual and semi-automatic segmentations were performed by two examiners on a Lenovo Yoga 2 Pro laptop with an Intel Core i7-4500U CPU @ 2.40 GHz, 8 GB RAM and Windows 8.1, 64 bit installed. The two examiners have over 10 years-experience in performing and interpreting US images and performed over 20,000 ultrasound examinations. Besides, the examiners did not know the images, nor have they worked with the algorithm before. In a first session, the examiners manually outlined all metastasis. In addition, they draw the largest diameter (diameter a) and an additional shorter diameter 90 degree related to the first one (diameter b). Time of measurement per metastasis was recorded. In a second session, five weeks later (to decrease bias due to memory effect), the examiners performed the semi-automatic segmentations. Note that the segmentation times have been recorded for every metastasis and the order of the metastasis was randomly redistributed in the second session (semi-automatic segmentation). However, the examiners were instructed in the semi-automatic segmentation with an additional set of ten images of pancreatic cancer liver metastases.

*C. Algorithm*

The segmentation algorithm has been integrated into the MeVisLab platform, which we used already for various medical applications [31]-[44] and has initially been developed and presented on a limited dataset of liver ultrasound images [45]-[48]. Summarized, the interactive segmentation algorithm needs a user-defined seed point inside the metastasis. Then, the algorithm automatically analyzes the gray values around this seed point like shown in other segmentation tasks [49]-[52], estimating an average gray value of the liver lesion. This makes the algorithm insensitive to different echopattern of homogenous livermasses in ultrasound B-mode, which can appear hyperechoic (brighter), isoechoic (similar) or hypoechoic (darker), when compared to the surrounding liver tissue. Besides, this course of action does not require any parameter changes for different acquisitions. Afterwards, the algorithm generates a specific graph, based on a circular template and performs a minimal s-t-cut. The result of the Min Cut is presented to the user in real-time as a contour outline around the liver lesion. This allows the user to drag the seed point around (inside the lesion) to improve the result until a satisfying segmentation is reached [53]. However, for difficult cases, the user can place additional seeds on the border of the metastasis, which restrict and support the interactive segmentation and was worked already in other structures [54]-[57]. Figure 1 presents the high-level workflow for the semi-automatic segmentation using the so-called US-Cut approach.

| | DSC (%) | | HD (Voxel) | | Difference Diameter a (mm) | | Difference Diameter b (mm) | |
|---|---|---|---|---|---|---|---|---|
| | *Median* | *MAD* | *Median* | *MAD* | *Median* | *MAD* | *Median* | *MAD* |
| **Examiner 1 (n=92)** | 84 | 6 | 9 | 4 | 2 | 2 | 1 | 1 |
| **Examiner 2 (n=94)** | 82 | 7 | 10 | 7 | 3 | 2 | 2 | 2 |

Table. 1. Summary results of the manual and satisfied semi-automatic measurements for two Internal Medicine specialists (examiner), with Dice Similarity Coefficient (DSC), Hausdorff distance (HD), standard deviation (SAD) and mean absolute deviation (MAD).

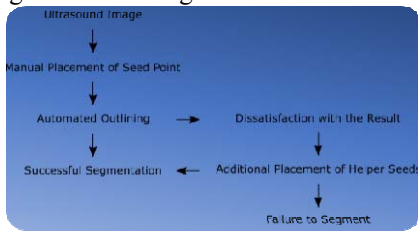

Fig. 1. Workflow of the semi-automatic segmentation approach.

III. RESULTS

For this study, 105 cases (77 ultrasound images of 46 patients diagnosed with PDAC and 28 images of 10 patients with neuroendocrine pancreatic neoplasia) have been used for the evaluation. The median of the maximal lesion diameter was 20 mm and all images displayed liver metastases without overlaying of marker or text. Table 1 presents the summary results of the manual and satisfied semi-automatic measurements for two Internal Medicine specialists (examiner), with the Dice Similarity Coefficient (DSC) [58], the Hausdorff distance (HD) [59], standard deviation (SAD) and mean absolute deviation (MAD). Overall, the examiners were satisfied with the algorithmic segmentation results in 92 (88%) and 94 (90%) cases with the result using the algorithm. However, in seven identical cases, both examiners were not satisfied with the algorithmic segmentation result, even with the support of helper seeds. The Dice score between the manual and semi-automatic segmentations was calculated for each examiner and yielded to a median DSC of 85% for examiner 1 and 82% for examiner 2. In addition, the median HD was calculated, which yielded to a low median distance of 9 pixels for examiner 1 and 10 pixel for examiner 2. Furthermore, the accuracy of the semi-automatic segmentation was analyzed, via the largest diameter (diameter a) with an additional shorter diameter (diameter b) that is 90 degree related to the first one. These were manually drawn by the examiner in addition to the metastasis outline. For the semi-automatic segmentation, these two diameters were calculated by the algorithm automatically based on the segmentation result. The diameter comparisons (manual / semi-automatic) yielded to a median difference of only 2 mm for diameter a and 1 mm for diameter b for examiner 1, and 3 mm for diameter a and 2 mm for diameter b for examiner 2. For a pure manual segmentation, examiner 1 (n=92) needed median 17.2 seconds, compared to examiner 2 (n=94), who needed median 10.2 seconds. In contrast, a semi-automatic segmentation was in median 7.7 seconds faster when performed by examiner 1 and 2.0 seconds faster when performed by examiner 2.

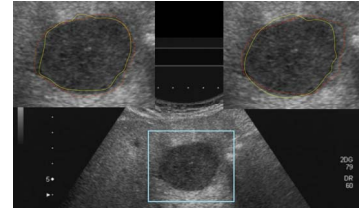

Fig. 2. Segmentation Example marked as adequate by both examiners. Upper left: Segmentation results of examiner 1. Upper right: Segmentation results of examiner 2. Red represent the manual segmentations and yellow represent the semi-automatic segmentations.

For visual inspection, Figure 2 and Figure 3 (left) present segmentation examples marked as adequate by both examiners. Finally, Figure 3 (right) presents a case marked as inadequate by both examiners.

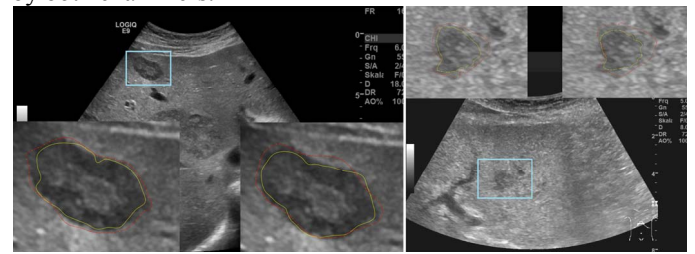

Fig. 3. Segmentation Example marked as adequate (left) and inadequate (right) by both examiners. Lower/Upper left: Segmentation results of examiner 1. Lower/Upper right: Segmentation results of examiner 2. Red represent the manual segmentations and yellow represent the semi-automatic segmentations.

IV. CONCLUSION

In this study, we presented the in-depth assessment of an interactive graph-based approach for the segmentation for pancreatic metastasis in ultrasound acquisitions of the liver. In doing so, two Internal Medicine specialists outlined over one hundred metastases of pancreatic cancer patients manually and with algorithmic support. Besides, the examiners had only a very short training phase of a few minutes to learn the usage of the algorithm. Albeit these challenges, the algorithm performed very well, shortened the segmentation time and could achieve a Dice score of over eighty percent. In addition, the algorithm achieved very small diameter deviations of only one to three millimeters and the examiners were satisfied with the segmentation results in almost ninety percent of the cases. To sum up, highlights of the proposed contribution are as follows: In-depth evaluation of an interactive graph-based approach; Manual outlining of over one hundred datasets by two Internal Medicine specialists; Algorithmic segmentation of the datasets by the two Internal Medicine specialists; Evaluation via Dice Similarity Coefficient, Hausdorff Distance, lesion diameters and segmentation time; Providing data to the research community. Despite the fact that the study presents encouraging results for a practical application of the algorithm in the clinical routine, there are some limitations that need to be addressed: The study had to be performed retrospectively, because of a missing direct access to the video-output of the ultrasound machine and the inclusion of more than one hundred images of different metastasis. So far, we studied only pancreatic cancer patients, because those are often staged during palliative treatment using abdominal ultrasound combined with endoscopic ultrasound in our clinic.

There are several areas for future work, like – as mentioned before – the direct streaming of the images from the US machines into our software for an immediate segmentation. Further, the enhancement and evaluation of our segmentation method for 3D US images [60]. Finally, providing the segmentation results in an Augmented Reality [61] or Virtual Reality [62] system.

ACKNOWLEDGMENT

The work received funding from BioTechMed-Graz and the 6[th] Call of the Initial Funding Program from the Research & Technology House (F&T-Haus) at the TU Graz (PI: Jan Egger). Dr. Xiaojun Chen received support from Foundation of Science and Technology Commission of Shanghai Municipality (14441901002, 15510722200, 16441908400). Anonymized raw data can be used for own research purposes as long as our work is cited at ResearchGate:

https://www.researchgate.net/publication/307907688_Ultrasound_Liver_Tumor_Datasets

Finally, videos demonstrating the interactive graph-based segmentation are available under the following YouTube channel: https://www.youtube.com/c/JanEgger/videos